\documentclass[letterpaper, 10 pt, conference]{ieeeconf}  % Comment this line out if you need a4paper

\IEEEoverridecommandlockouts                              % This command is only needed if 
\usepackage[utf8]{inputenc}
\usepackage[english]{babel}
\usepackage{graphicx}
\usepackage{amsmath,amssymb}
\usepackage{pifont}
\usepackage{amsfonts}
\usepackage{upgreek}
\usepackage{color}
\usepackage{marginnote}
\usepackage[ruled,vlined]{algorithm2e}
\usepackage{multirow}
\usepackage{paralist}
\usepackage{wrapfig}
\usepackage{xcolor}
\usepackage{hyperref}

\let\labelindent\relax
\usepackage{amsthm}
\usepackage{environ}
\usepackage[normalem]{ulem}
\usepackage{enumitem}
\usepackage{subcaption}
\usepackage{url}
\usepackage{hyperref}

\newtheorem{theorem}{Theorem}

\newtheorem{definition}{Definition}
\newtheorem{proposition}{Proposition}

\newtheorem*{remark}{Remark}

\usepackage{fancyvrb}
\usepackage{listings}
\newcommand{\reals}{\mathbb{R}}

\newcommand{\rp}{\reals^{\geq 0}}
\newcommand{\X}{\mathcal{X}}
\newcommand{\Y}{\mathcal{Y}}

\newcommand{\T}{\intercal}
\newcommand{\D}{\mathcal{D}}

\newcommand{\W}{\mathcal{W}}

\newcommand{\A}{\mathcal{A}}

\newcommand{\E}{\mathop{\mathbb{E}}}

\newcommand{\ellipsoid}[1]{\mathcal{E}\left(#1\right)}

\newcommand{\indicator}{\mathbb{I}}
\newcommand{\indicatorP}[1]{\indicator_{\rp}\left(#1\right)}

\newcommand{\br}[1]{\left(#1\right)}

\newcommand{\pa}{\A}

\title{\LARGE \bf
Learning-based Inverse Perception Contracts and Applications}

\author{Dawei Sun, Benjamin C. Yang and Sayan Mitra% <-this % stops a space
\thanks{This work is supported in part by NASA ULI grant NH 00723829 and the Boeing Company. The authors are with coordinated Science Laboratory, University of Illinois at Urbana-Champaign
{\tt\small \{daweis2, bcyang2, mitras\}@illinois.edu}}%
}

\begin{document}
\maketitle
\thispagestyle{empty}
\pagestyle{empty}
\begin{abstract}
Perception modules are integral in many modern autonomous systems, but their accuracy can be subject to the vagaries of the environment. In this paper, we propose a learning-based approach that can automatically characterize the error of a perception module from data and use this for safe control. The proposed approach constructs an {\em inverse perception contract (IPC)\/} which generates a set that contains the ground-truth value that is being estimated by the perception module, with high probability.
We apply the proposed approach to study a vision pipeline deployed on a quadcopter. With the proposed approach, we successfully constructed an IPC for the vision pipeline. We then designed a control algorithm that utilizes the learned IPC, with the goal of landing the quadcopter safely on a landing pad. Experiments show that with the learned IPC, the control algorithm safely landed the quadcopter despite the error from the perception module, while the baseline algorithm without using the learned IPC failed to do so.
\end{abstract}

\section{Introduction}
Perception plays a crucial role in making modern autonomous systems react and adapt in unstructured environments. In the past decade, we have seen perception algorithms invented to perform a variety of  tasks such as  detection and classification of obstacles~\cite{girshick2015fast}, pose estimation~\cite{xiang2017posecnn}, and localization~\cite{mur2015orb}.

However, any interpretation of signals from the real world has to deal with noise, which is sometimes poorly understood and characterized.
As the output of the perception module drives downstream decisions or control modules in an autonomous system, the errors in perception may result in unwanted and even catastrophic actions. Therefore, the problem of designing reliable perception-based control algorithms for safety-critical systems has become an active research topic, and attempts have been made in recent works, for example~\cite{dean2021guaranteeing, dawson2022learning}.

One way to design reliable perception-based controllers is to first characterize the perception error and then design controllers that are robust to the perception error. In a series of recent works~\cite{hsieh2022verifying, hsieh2023perception}, the notion of {\em perception contracts (PC)\/} has been proposed to address this problem. A PC characterizes the error of perception modules in a way that can be useful for proving system-level invariants. The authors have shown that such perception contracts can be automatically constructed from data and they used PCs to verify several vision-based lane-keeping systems~\cite{hsieh2022verifying, hsieh2023perception}. In this paper, we aim to solve the controller synthesis problem instead. Inspired by the idea of PCs, we proposed inverse perception contracts (IPCs) that map a perceived value to a perception error, which characterizes the uncertainty of the perception result\footnote{In the earlier works, the PC instead maps the ground truth values to perception error.}.
Such an IPC can then be used by the downstream controller and decision-making modules to compensate for the uncertainty of the perception, which contributes to enhancing the safety and robustness of the autonomous system.

In this paper, we consider perception modules that aim to estimate the value of a specific quantity, for example, the position of an obstacle. Let $y$ be the ground-truth value of this quantity. Most perception algorithms only provide a single-point estimate $\hat{y}$, which can be seen as a noisy version of $y$. However, in order to design robust controllers, the uncertainty of the estimate $\hat{y}$ is also needed. To this end, we construct an IPC for the perception module, which is simply a mapping from the perceived value $\hat{y}$ to a set that contains the ground-truth value $y$ with a high probability. This containment should hold even under environmental variations. We model the IPC using a neural network and carefully design loss functions that balance the error and conservativeness of the IPC. The only requirement for applying the proposed approach is a data set that consists of pairs of estimate $\hat{y}$ and ground truth $y$. Once trained on the data set, the IPC can be executed online and quickly compute the uncertainty of any estimate $\hat{y}$. The computed uncertainty can then be used by downstream modules to compute safe control signals.

We evaluated the proposed approach on a quadcopter equipped with a camera. The perception module of interest is a program that aims to estimate the pose of a landing pad from camera images. By applying the proposed approach, we successfully constructed an accurate IPC for the perception module. Experimental results show that the error of the learned IPC was below $0.2\%$ on the testing set. In order to study the conservativeness of the learned IPC, we tuned some parameters of the perception module to change the characteristics of the perception error. We measured and reported the error of the IPC under these unseen perception parameters. Results show that the output of the learned IPC is tight and adheres to the perception parameter under which it was trained. Then, in order to show the IPC's capability of being used by downstream modules, we design a robust control algorithm that utilizes the learned IPC and aims to safely land the quadcopter on the landing pad. Experiments show that with the learned IPC, the quadcopter can safely complete the landing task despite the significant perception error.

In summary, our contribution is threefold.
\begin{itemize}
    \item We proposed an approach that can automatically learn an inverse perception contract from data;
    \item We demonstrated that the learned inverse perception contract can be used by downstream modules to compensate for the error of the upstream perception module;
    \item We evaluated the proposed approach in a real-world application.
\end{itemize}

\section{Related work}
\label{sec:related}
\paragraph{Perception-based control}
As new types of sensors emerge, the problem of integrating these sensors and the corresponding perception modules into the control pipeline has attracted interest. In recent works~\cite{lin2018autonomous,loianno2016estimation, tang2018aggressive}, perception-based controllers have been studied to enable aggressive control for quadcopters. Further, data-driven approaches have been developed by the machine learning community. For example, imitation learning~\cite{codevilla2018end} and reinforcement learning~\cite{Sadeghi-RSS-17} have been used to learn vision-based control policies. However, those approaches cannot provide any guarantees for the synthesized controller, which limits their application in safety-critical applications.

As autonomous system enters more and more safety-critical domains, designing perception-based control with formal safety guarantees has become a key problem in robotics research. In a series of recent works by Dean et al., the authors studied the robustness guarantees of perception-based control algorithms under simple perception noises: In~\cite{dean2020robust}, the authors proposed a perception-based controller synthesis approach for linear systems and provided a theoretical analysis. In~\cite{dean2021guaranteeing}, the authors proposed robust barrier functions that provide an approach for synthesizing safety-critical controllers under uncertainty of the state. More complex perception modules have also been studied. In~\cite{dawson2022learning}, the authors studied the perception-based control synthesis problem for a vehicle with a Lidar. The authors use neural networks to learn a control Lyapunov function (CLF) and a control barrier function (CBF) in the observation space, which enable safe navigation in an environment with unknown obstacles. In~\cite{chou2022safe}, a perception module is learned with neural networks from data. Instead of a single state, the perception module outputs a set of states. Then, the authors apply contraction theory and robust motion planning algorithms to synthesize control that is robust to the perception error. Our approach is similar to the one proposed in~\cite{chou2022safe} in the sense that they both introduce a high-confidence set for the perception results. In contrast, our approach can be applied to construct such high-confidence sets for an existing perception module, while in~\cite{chou2022safe}, the authors have to construct these sets while designing the perception module.

\paragraph{Analysis of systems with black-box modules} The proposed approach is also related to works concerning the analysis of systems with black-box modules. As the complexity of autonomous systems increases, it becomes impossible to analyze them exactly. Many modules in modern autonomous systems are black boxes, for example, modules based on neural networks.
Data-driven approaches have been applied to analyze those systems instead. The approach of constructing an inverse perception contract from data is mainly adapted from the reachability analysis approach proposed in~\cite{sun2022neureach}, where machine learning is used to analyze the reachability of black-box systems.
VerifAI~\cite{dreossi2019verifai} provides a complete framework for analyzing autonomous systems with ML modules in the loop.
The notion of perception contracts was originally proposed in~\cite{hsieh2022verifying}, where the authors study safe abstractions of ML-based perception modules. \cite{hsieh2022verifying} is focused on verifying properties of an existing system and thus concerned with predicting the estimate $\hat{y}$ given the ground truth $y$. In contrast, this paper is focused on the synthesis problem and thus aims to predict the ground truth $y$ given the estimate $\hat{y}$. Thus, we adopted the idea of perception contracts but used its inverse instead.

\section{Learning inverse perception contracts}
\begin{figure}
    \centering
    \includegraphics[width=\linewidth]{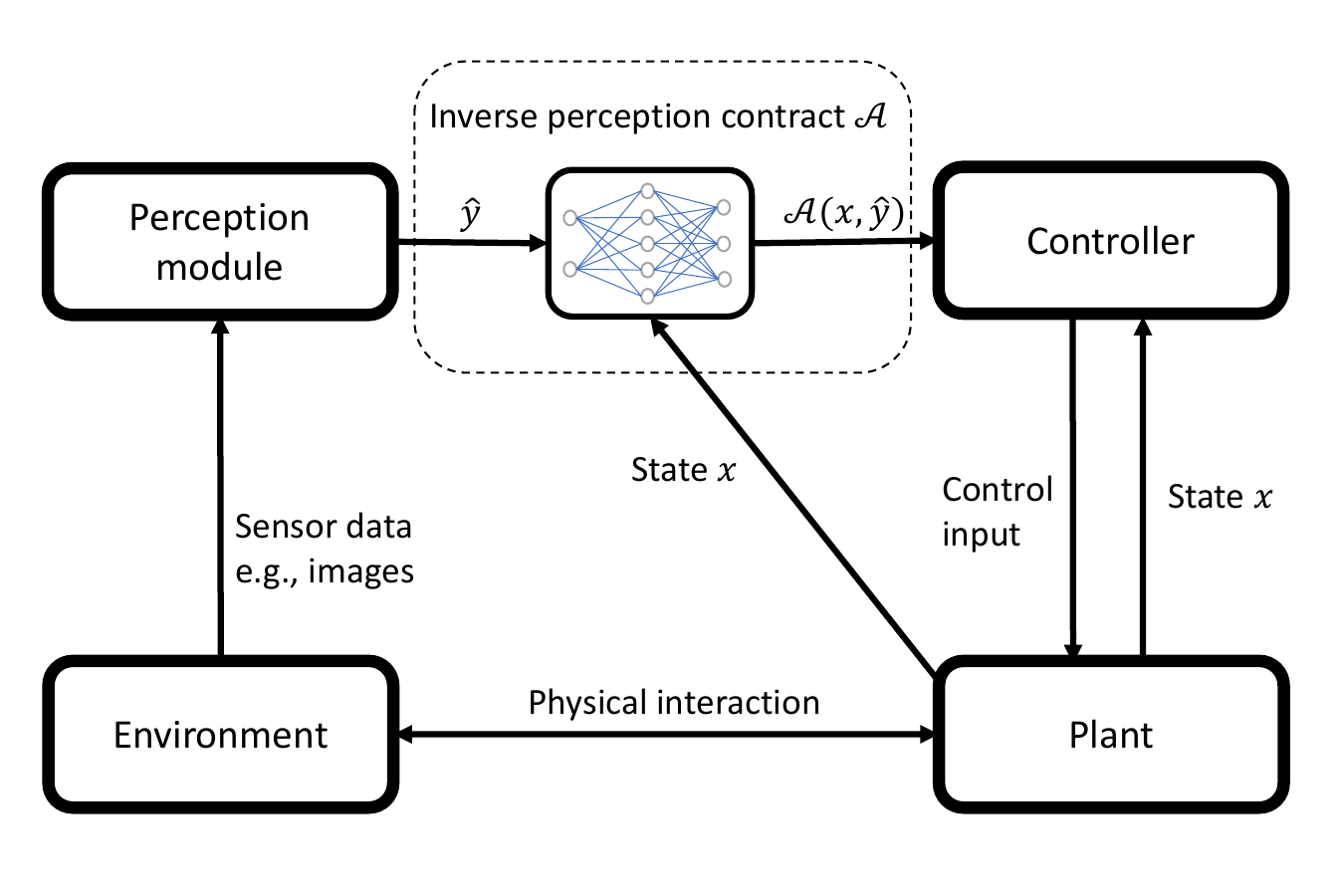}
    \caption{An autonomous system that entails a perception module and the proposed inverse perception contract.}
    \label{fig:perception_control}
\end{figure}
In this paper, we consider general control systems that entail perception modules as shown in Figure~\ref{fig:perception_control}. The state of the system is denoted by $x \in \X$, where $\X$ is the state space of the system. For example, if the system is a quadcopter, the state vector $x$ may include the position, velocity, and attitude of the quadcopter.

For a control system that interacts with the real world, access to only the state $x$ of the system itself is not enough. It is very common that the value of an external quantity  $y \in \Y \subseteq \reals^n$ is also needed in order to complete a task. For example, for a quadcopter that is asked to land, $y$ can be the position of the landing pad. However, access to the ground-truth value of $y$ is often unavailable. To this end, a {\em perception module} is used to estimate the value of $y$.

As shown in Figure~\ref{fig:perception_control}, we assume that a perception module is given {\it a priori}. The perception module aims at estimating the value of $y$. We denote the output of the perception module by $\hat{y}$ and call it the perceived value.
Due to the complexity and uncertainty of the real world, the perceived value does not always coincide with the ground-truth value, and there is a non-zero perception error $e = \hat{y} - y$.

\subsection{Inverse perception contracts}
In order to design a robust controller that utilizes the perception results, one has to analyze the perception module and characterize the perception error. That is, given the state $x$ and the perceived value $\hat{y}$, one has to be able to know some characteristics of the perception error $e$, for example, the maximum of $\|e\|$. However, it is a challenging task. Although there is a relationship between the state $x$, the ground-truth value $y$, and the perceived value $\hat{y}$, it is usually not possible to find a closed-form expression of this relationship. Inspired by~\cite{hsieh2022verifying}, we propose to use {\em inverse perception contracts} as a unified way of characterizing perception errors. Inverse perception contracts then serve as an interface between the perception module and the controller synthesis algorithm. Perception modules and controller synthesis algorithms that follow the same interface can be freely paired.

In general, the relationship between the state $x$, the ground-truth value $y$, and the perceived value $\hat{y}$ is not deterministic. Thus, we assume that $x$, $y$ and $\hat{y}$ conform to a joint distribution $\D$, i.e., $x, y, \hat{y} \sim \D$. We aim at finding a way to recover $y$ from $x$ and $\hat{y}$. To this end, we define an {\em inverse perception contract} as a mapping $\pa : \X \times \Y \to 2^\Y$, $x, \hat{y} \mapsto \pa(x, \hat{y})$, where $2^\Y$ is the power set of $\Y$. Ideally, given the state $x$ and the perceived value $\hat{y}$, the output of the inverse perception contract $\pa(x, \hat{y})$ should contain the ground-truth value $y$, i.e., $y \in \pa(x, \hat{y})$. However, such a requirement for the inverse perception contract is too strong, given the fact that the relationship between $x$, $y$, and $\hat{y}$ is stochastic. Instead, we allow approximately correct inverse perception contracts, and the error of an inverse perception contract $\pa$ is defined as follows.
\begin{definition}[Error of an inverse perception contract]
The error of an inverse perception contract $\pa : \X \times \Y \to 2^\Y$ is
\begin{equation}
\Pr_{x, y, \hat{y} \sim \D} \br{y \notin \pa(x, \hat{y})}.
\end{equation}
\end{definition}

In this paper, we study the problem of constructing an inverse perception contract of a given perception module. Next, we will propose an approach to this problem, where we model the inverse perception contract using a neural network.

\subsection{Learning inverse perception contracts from data}
In this section, we propose an algorithm for learning an inverse perception contract from data. The inverse perception contract is modeled with a neural network. By definition, the output of an inverse perception contract is a set.
In order to use a neural network to model an inverse perception contract, we need to assume that the output of the inverse perception contract is from a finite-dimensional domain, i.e., the elements of this domain can be represented with finite-dimensional vectors. In this paper, we adopt ellipsoids as the domain. That is, the output of the inverse perception contract is always an ellipsoid.

In theory, any finite-dimensional domain such as hyper-rectangles or zonotopes could be used as the output of the inverse perception contract. We adopt ellipsoids due to their smoothness and simple representations. An ellipsoid in $\reals^n$ is defined by a center $c \in \reals^n$ and a non-singular matrix $C \in \reals^{n \times n}$ and is denoted by $\ellipsoid{c, C} := \{x \in \reals^n : \|C(x - c)\|_2 \leq 1\}$. Thus, in order to model an inverse perception contract whose output is an ellipsoid, we only need two parametric functions $c_{\theta} : \X \times \Y \to \reals^n$ and $C_{\theta} : \X \times \Y \to \reals^{n \times n}$ with parameters $\theta \in \W$. Given a state $x$ and a perceived value $\hat{y}$, $c_\theta$ and $C_\theta$ defines an ellipsoid with center $c_\theta(x, \hat{y})$ and shape $C_\theta(x, \hat{y})$. Therefore, the parametric inverse perception contract is as follows
\begin{equation}
\pa_\theta(x, \hat{y}) := \ellipsoid{c_\theta(x, \hat{y}), C_\theta(x, \hat{y})}.
\end{equation}
As will be shown in Section~\ref{sec:learning}, $c_\theta$ and $C_\theta$ are actually modeled as two heads of the same neural network.

For the sake of simplicity, we denote $X = (x, y, \hat{y})$ and define a helper function
\begin{equation}
g_{\theta}(X) := \left\|C_{\theta}(x, \hat{y}) \left(y - c_\theta(x, \hat{y})\right)\right\|_2.
\end{equation}
Clearly, $g_\theta(X) \leq 1$ if and only if $y \in \pa_{\theta}(x, \hat{y})$.

We aim at designing a learning algorithm to find a $\theta$ that minimizes the error of the parametric inverse perception contract $\pa_\theta$. In other words, we want to minimize the following loss function.
\begin{equation}
L(\theta) := \E\limits_{X \sim \D}\left[\indicatorP{g_{\theta}(X) - 1}\right],
\end{equation}
where $\indicatorP{\cdot}$ is the indicator function, i.e., $\indicatorP{x} = 1$ if $x > 0$, and $\indicatorP{x} = 0$ otherwise.

In practice, the closed-form expression of the underlying distribution $\D$ is not available, and one can only sample from $\D$. Thus, the above loss function cannot be minimized directly since the expectation over $\D$ cannot be computed exactly. Therefore, we resort to empirical risk minimization as follows. First, a training set $S = \{X_i\}_{i=1}^{N}$ is constructed, where the samples $X_i = (x^{(i)}, y^{(i)}, \hat{y}^{(i)})$ are independently drawn from the data distribution $\D$. Then, the {\em empirical loss\/} on $S$ is defined as follows.
\begin{equation}
L_{ERM}(\theta) = \frac{1}{N}\sum_{i=1}^{N} \ell\left(g_{\theta}(X_i) - 1\right),
\label{eq:ERM_loss}
\end{equation}
where $\ell(x) := \max\{0, \frac{x}{\alpha} + 1\}$ is the hinge loss function with the hyper-parameter $\alpha>0$, which can be seen as a soft proxy for the indicator function.

In addition to minimizing the error of the inverse perception contract, we also have to penalize its conservativeness. Otherwise, we might get trivial inverse perception contracts with extremely low error. For example, the error of the inverse perception contract $\pa(x, \hat{y}) \equiv \Y$ is $0$, but such an inverse perception contract is useless in the design of robust controllers. Thus, the volume of the ellipsoid must be penalized. Inspired by~\cite{pmlr-v120-devonport20a}, we use $-\log(|C^\T C|)$ as a proxy of the volume of an ellipsoid $\ellipsoid{c, C}$, and the following regularization term is added to penalize conservative outputs.
\begin{equation}
L_{REG}(\theta) = -\frac{1}{N}\sum_{i=1}^{N} \log\Big(\left|C_{\theta}(x^{(i)}, \hat{y}^{(i)})^\T C_{\theta}(x^{(i)}, \hat{y}^{(i)})\right|\Big).
\end{equation}

In practice, we solve the following optimization problem with stochastic gradient descent.
\begin{equation}
\hat{\theta} = \arg\min_{\theta} L_{ERM}(\theta) + \lambda L_{REG}(\theta),
\label{eq:ERM_opt}
\end{equation}
where $\lambda$ is a hyper-parameter balancing the error and conservativeness of the inverse perception contract.

\subsection{Probabilistic correctness of the inverse perception contract}
\label{sec:analysis}
With standard techniques from statistical learning theory, we can derive the following theorem which gives an upper bound on the error of the learned inverse perception contract.
\begin{theorem}
For any $\epsilon>0$, and a random training set $S$ with $N$ i.i.d. samples, with probability at least $1-2 \exp(-2N\epsilon^2)$, the following inequality holds,
\begin{multline}
\E\limits_{X \sim \D}\left[\indicatorP{g_{\hat{\theta}}(X) - 1}\right]\\ \leq \frac{1}{N}\sum_{i=1}^{N} \tilde{\ell}(g_{\hat{\theta}}(X_i) - 1) + \frac{12}{\alpha}L_{g}\sqrt{\frac{p}{N}} + \epsilon,
\label{eq:pac}
\end{multline}
where $p$ is the number of scalar parameters of the neural network, and $\tilde{\ell}(\cdot) = \min\{1, \ell(\cdot)\}$ is the truncated hinge loss, and $L_g$ is the Lipschitz constant of $g_{\theta}$ w.r.t. $\theta$.
\label{thm:pac_bound}
\end{theorem}

\begin{remark}
Theorem~\ref{thm:pac_bound} shows that by controlling $\epsilon$ and $N$, the actual error $\E_{X \sim \D}\left[\indicatorP{g_{\hat{\theta}}(X) - 1}\right]$ can be made arbitrarily close to the empirical loss $\frac{1}{N}\sum_{i=1}^{N} \tilde{\ell}(g_{\hat{\theta}}(X_i) - 1)$, with arbitrarily high probability.
Furthermore, the empirical loss can be made very small in practice due to the high capacity of the neural network. Of course, there is no free lunch in general. In order to drive the empirical loss to $0$, we might have to increase the number of parameters, which in turn increases the term $\frac{12}{\alpha}L_{g}\sqrt{\frac{p}{N}}$.
\end{remark}

\section{Application study: safe quadcopter landing}
In this section, we evaluate the approach proposed in the last section in a real-world application, namely, safe quadcopter landing.
\begin{figure}
    \centering
    \begin{subfigure}[b]{0.42\linewidth}
        \includegraphics[width=\textwidth]{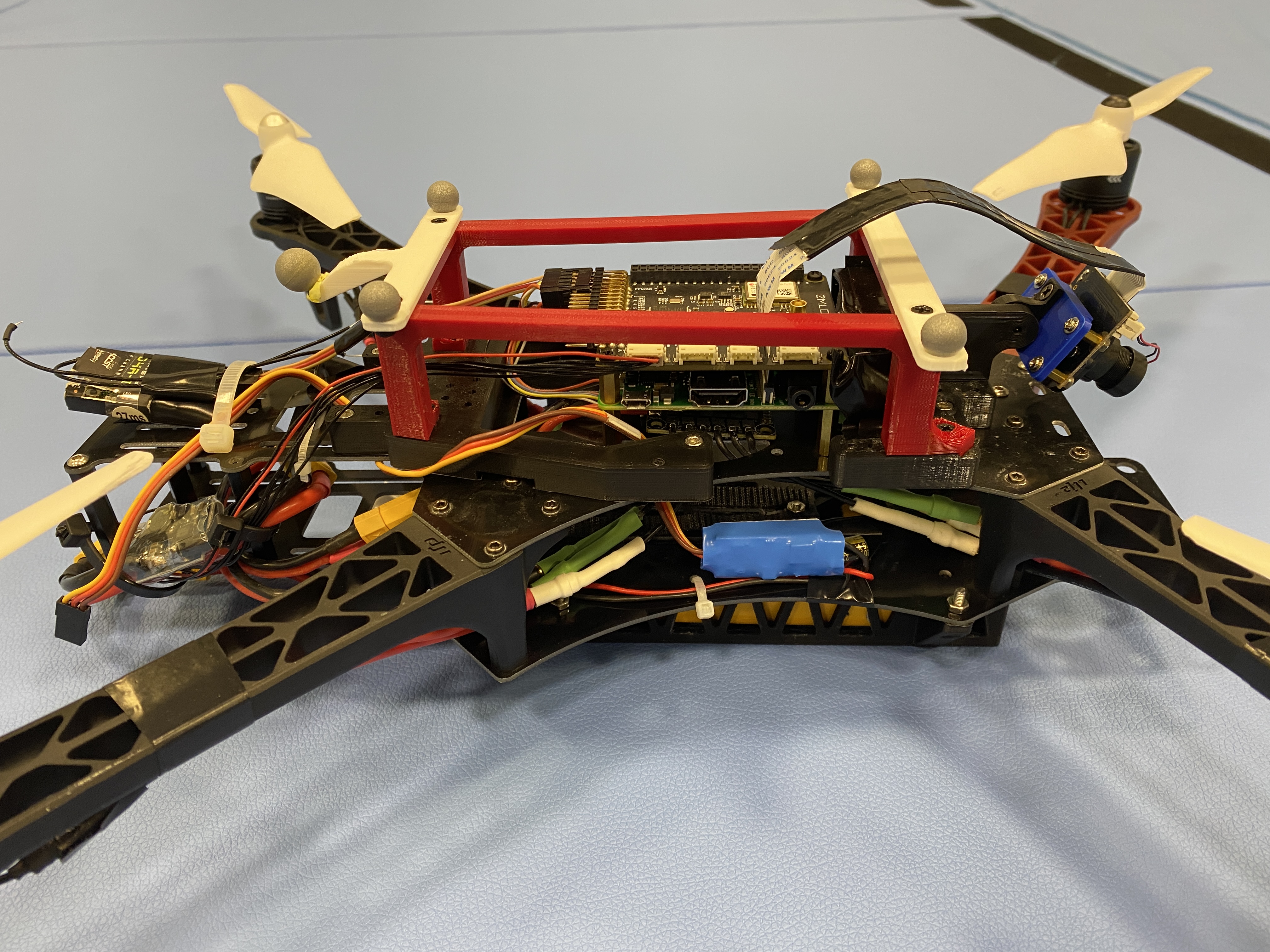}
        \caption{}
        \label{fig:quadcopter}
    \end{subfigure}
    \hfill
    \begin{subfigure}[b]{0.45\linewidth}
        \includegraphics[width=\textwidth]{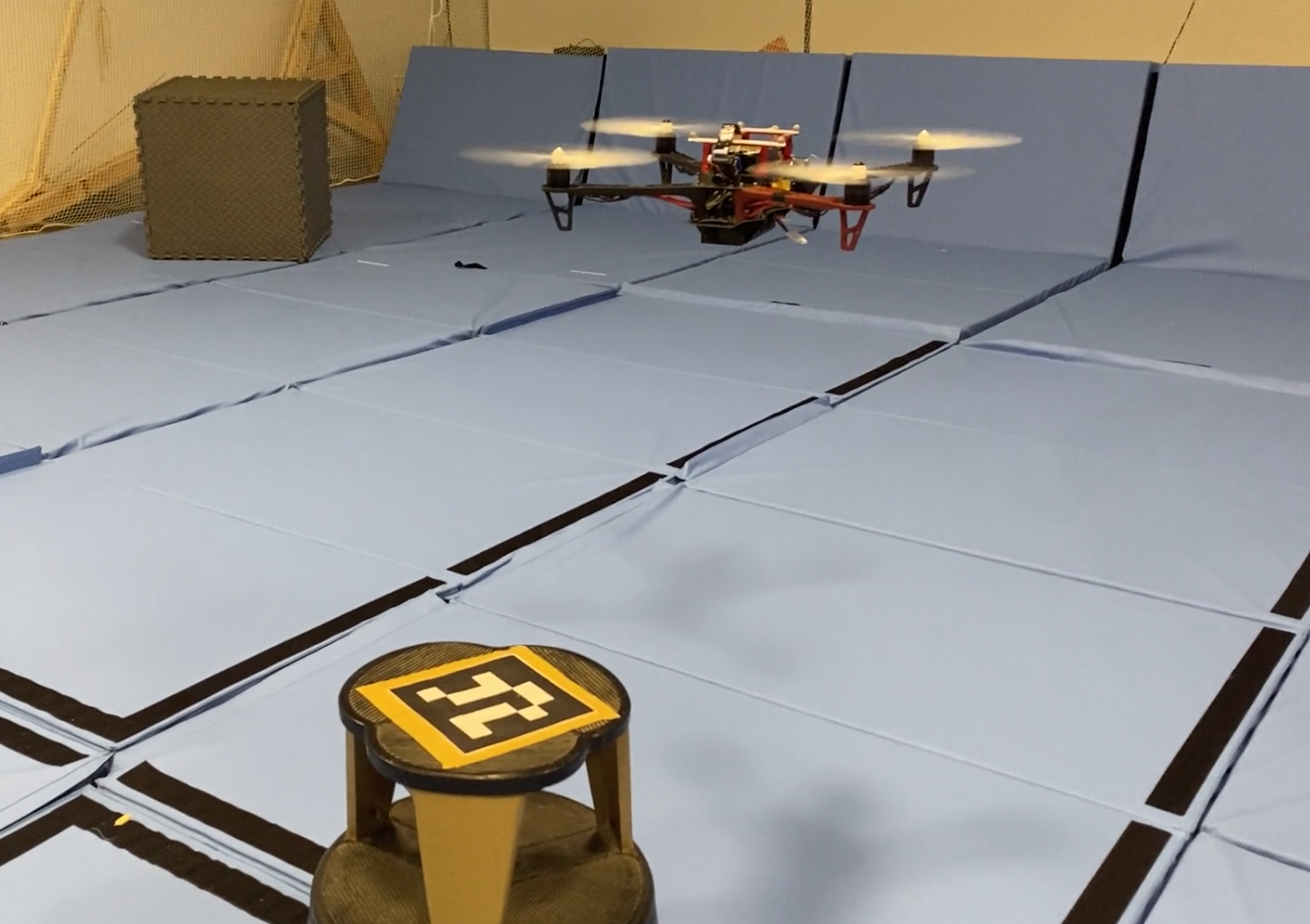}
        \caption{}
        \label{fig:landing_pad}
    \end{subfigure}
    \caption{The quadcopter and the landing pad.}
\end{figure}

\subsection{The safe landing problem}
\paragraph{The quadcopter} In this section, we consider the problem of designing an algorithm for safely landing the quadcopter as shown in Figure~\ref{fig:quadcopter}. The quadcopter is built based on a DJI\textsuperscript{\textregistered} F450 frame. The size of the quadcopter is $36$cm $\times$ $36$cm $\times$ $10$cm measured without propellers. A Raspberry Pi 3\textsuperscript{\textregistered} computer and a Navio2\textsuperscript{\textregistered} board\footnote{\url{https://navio2.hipi.io/}} are mounted on the quadcopter. The Navio2\textsuperscript{\textregistered} board provides essential sensors for quadcopter control such as inertial measurement units (IMU) and a barometric pressure sensor. A camera is connected to the Raspberry Pi computer. The camera can produce $320 \times 240$ images at a rate of $60$ frames per second. The state of the quadcopter is
\begin{equation}
x := [p_x, p_y, p_z, v_x, v_y, v_z, \phi_x, \phi_y, \phi_z, \omega_x, \omega_y, \omega_z],
\end{equation}
where $p_x$, $p_y$, and $p_z$ are the 3D position of the quadcopter and similarly, $v$, $\phi$ and $\omega$ are velocity, attitude and angular velocity respectively.

\paragraph{The workspace} The quadcopter is operating in a $6\text{m} \times 6\text{m} \times 3\text{m}$ workspace. Furthermore, a motion capture system (Vicon\textsuperscript{\textregistered}) is used for low-latency and high-accuracy localization of the quadcopter. The state $x$ estimated by the Vicon\textsuperscript{\textregistered} system is viewed as the ground-truth state of the quadcopter. As shown in Figure~\ref{fig:landing_pad}, a landing pad is placed on the ground of the workspace. The goal of the quadcopter is to land on the landing pad. Although the quadcopter has access to its state $x$ estimated by the Vicon\textsuperscript{\textregistered} system, the position of the landing pad is {\em unknown} to the quadcopter. We denote the ground-truth position of the landing pad by $y \in \reals^3$.

\paragraph{Coordinate frames} There are several coordinate frames that will be used in the experiments. They are the world frame $W$, the quadcopter frame $Q$, and the camera frame $C$. For the same quantity, we use subscripts to discriminate among its values under different coordinate frames. For example, $y_W$ is the coordinate of the landing pad in the world frame, while $y_C$ is that in the camera frame. When the subscript is omitted, it denotes the value in the world frame.

\paragraph{The vision-based perception module} In order to land on the landing pad, the quadcopter uses the camera to estimate the position $y_W$. To this end, we attach an ArUco marker~\cite{garrido2014automatic} to the landing pad. An ArUco marker is basically a QR-code-like image, which enables robust and fast pose estimation from images. By calling functions implemented in the OpenCV library~\cite{opencv_library}, we can estimate the position of the landing pad in the camera frame, and such an estimate is denoted by $\hat{y}_C$. Then, we transform $\hat{y}_C$ into the world frame $\hat{y}_W$ using the extrinsic parameter between the camera and the quadcopter, and the quadcopter's pose $x_W$ in the world frame. However, $\hat{y}_C$ and $x_W$ are measured by two different sensors (camera and Vicon) without synchronization, and this asynchronization contributes as the main source of the perception error.

\paragraph{The task specification} The goal of the quadcopter is to navigate to a box around the ground-truth position $y$ of the landing pad and then turn off the motors.
The box is defined as $\{y\} \oplus G$ where $G$ is a $0.1 m \times 0.1 m \times 0.05 m$ box defined as $G := [-0.05, 0.05] \times [-0.05, 0.05] \times [0, 0.05]$, and $\oplus$ is the Minkowski addition.
The quadcopter is said to violate the safety requirement if it stops the motors outside $\{y\} \oplus G$.

\subsection{Learning an inverse perception contract}
\label{sec:learning}
Next, we apply the proposed approach to learn an abstraction of the perception module. The most straightforward idea is to directly apply the proposed approach to learn an IPC as a function of $x$ and $\hat{y}$ in the world frame. However, by doing this, we bind the learned IPC to a specific transformation between the quadcopter and the camera. Such an IPC will not be usable after the transformation between the quadcopter and camera is changed, for example, when the camera direction is switched from forward to downward. To tackle this issue, we instead learn an IPC in the camera frame, i.e., $x_C$, $y_C$, and $\hat{y}_C$ will be used in learning. As the first step, we construct a data set.

\paragraph{Construction of the data set} During the process of collecting data, we fix the position of the landing pad and measure its ground-truth position $y$ beforehand. Then, we program the quadcopter to follow some predefined paths and record the state $x$ and the perceived value $\hat{y}$ on the fly. The pre-defined paths are designed such that by following it the quadcopter explores diverse positions with diverse velocities in the workspace. Please see the supplementary video for more details on collecting data. We then transform all the quantities into the camera frame and obtained the data set $\{(x_C^{(i)}, y_C^{(i)}, \hat{y}_C^{(i)})\}_{i=1}^{N}$ with $N = 5000$.

\paragraph{Machine learning setup}
We model the IPC using a three-layer neural network of which the hidden layers contain $64$ and $128$ neurons. The input of the neural network is the concatenation of the state $x_C$ and the perceived value $\hat{y}_C$. The output size of the neural network is $12$. The first three elements of the neural network's output are used as the center of the ellipsoid and the remaining nine elements are reshaped into a $3 \times 3$ matrix that describes the shape of the ellipsoid. We train the neural network for $20$ epochs using the Adam~\cite{kingma2014adam} optimizer. The learning rate is set to $0.001$.

\paragraph{Test of the learned inverse perception contract}
After training, we evaluate the learned IPC using real data and estimate its error. To do that, we manually control the quadcopter to fly in the workspace until the quadcopter finishes $\sim 300$ measurements of the landing pad. For each measurement, we check whether $y \in \pa(x, \hat{y})$ and compute an empirical error rate.
As mentioned before, the perception error mainly comes from the asynchronization between the camera and Vicon. By tuning the camera buffer size in the OpenCV library, we can change the extent to which the camera image lags from the Vicon measurements. For each buffer size configuration, we conducted the same test and reported the error of the learned IPC in Table~\ref{tab:buffer_size_acc}. The training data was collected at buffer size $= 1$. As can be seen from Table~\ref{tab:buffer_size_acc}, the error of the learned IPC is as low as $0.19 \%$ at buffer size $= 1$. On the other hand, the learned IPC adheres to the buffer size configuration under which it was trained. As buffer size increases, the error of the learned IPC drastically increases, which suggests that the output of the IPC is not conservative. Moreover, it takes only $\sim 3$ milliseconds on the Raspberry Pi computer to compute $\pa(x, \hat{y})$ for each query $x$ and $\hat{y}$, which enables it to be used in online control pipelines.

\begin{table}[tbp]
    \centering
    \caption{Error rate of the learned inverse perception contract under different camera buffer size configurations.}
    \begin{tabular}{|c|c|c|c|c|c|}
    \hline
    Buffer size & 1 & 2 & 3 & 4 & 5\\ \hline
    Error & $0.19\%$ & $0.93\%$ & $2.05\%$ & $4.07\%$ & $11.05\%$\\ \hline
    \end{tabular}
    \label{tab:buffer_size_acc}
\end{table}

\subsection{Utilizing the learned inverse perception contract}
\begin{figure}
    \centering
    \includegraphics[width=\linewidth]{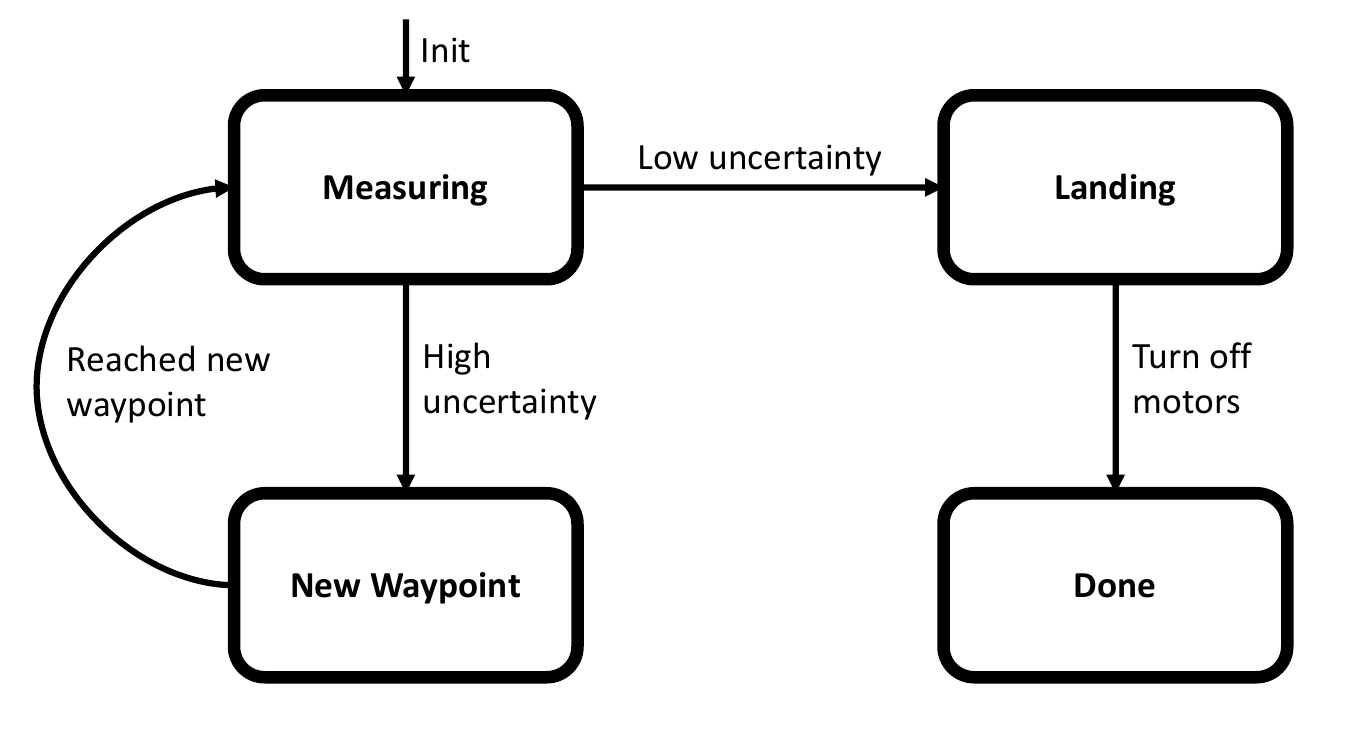}
    \caption{The state machine of the safe landing algorithm.}
    \label{fig:fsm}
\end{figure}

\begin{figure*}[tbp]
    \centering
    \begin{subfigure}[b]{0.31\textwidth}
        \includegraphics[width=\textwidth]{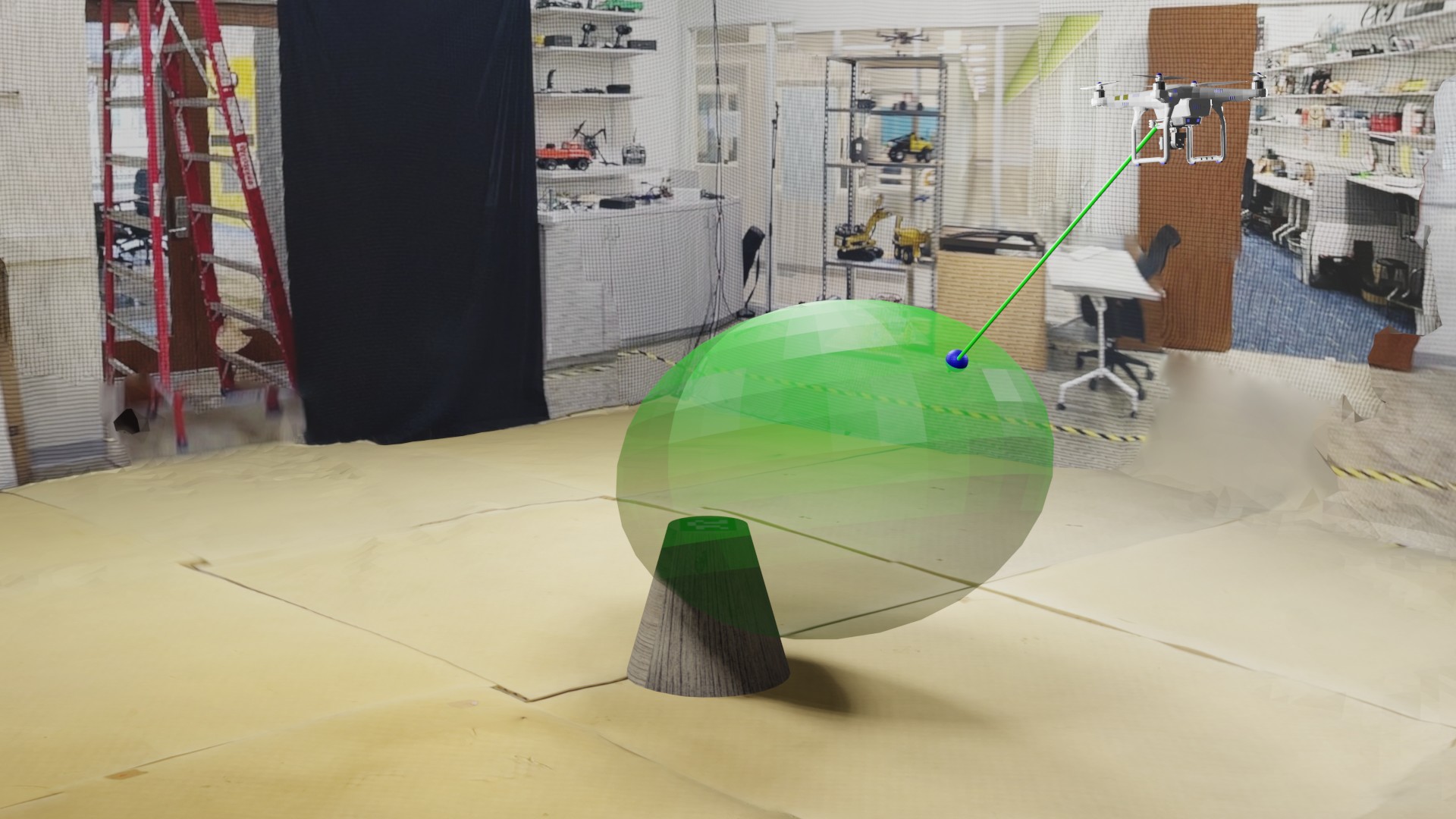}
        \caption{}
    \end{subfigure}
    \hfill
    \begin{subfigure}[b]{0.31\textwidth}
        \includegraphics[width=\textwidth]{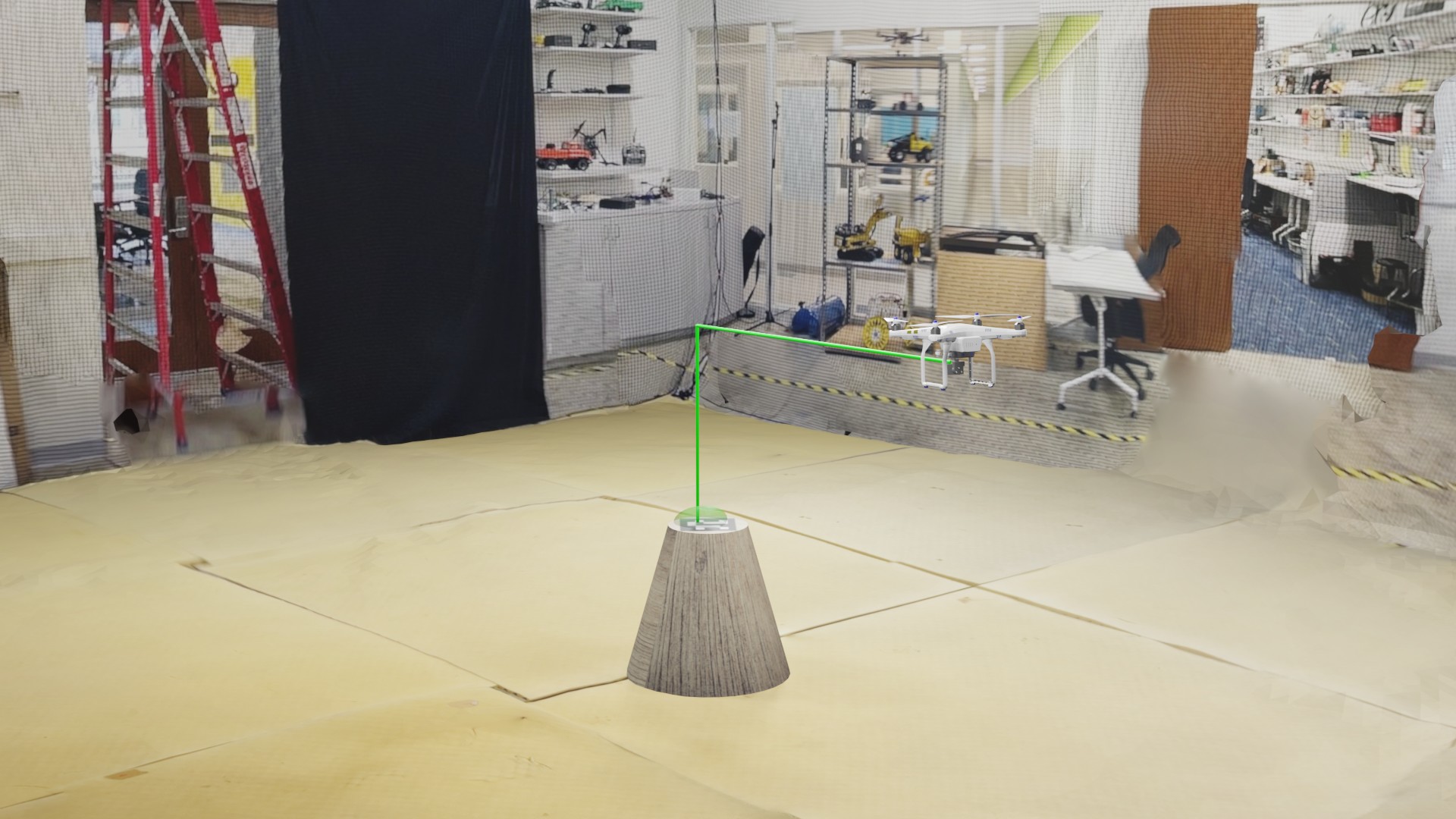}
        \caption{}
    \end{subfigure}
    \hfill
    \begin{subfigure}[b]{0.31\textwidth}
        \includegraphics[width=\textwidth]{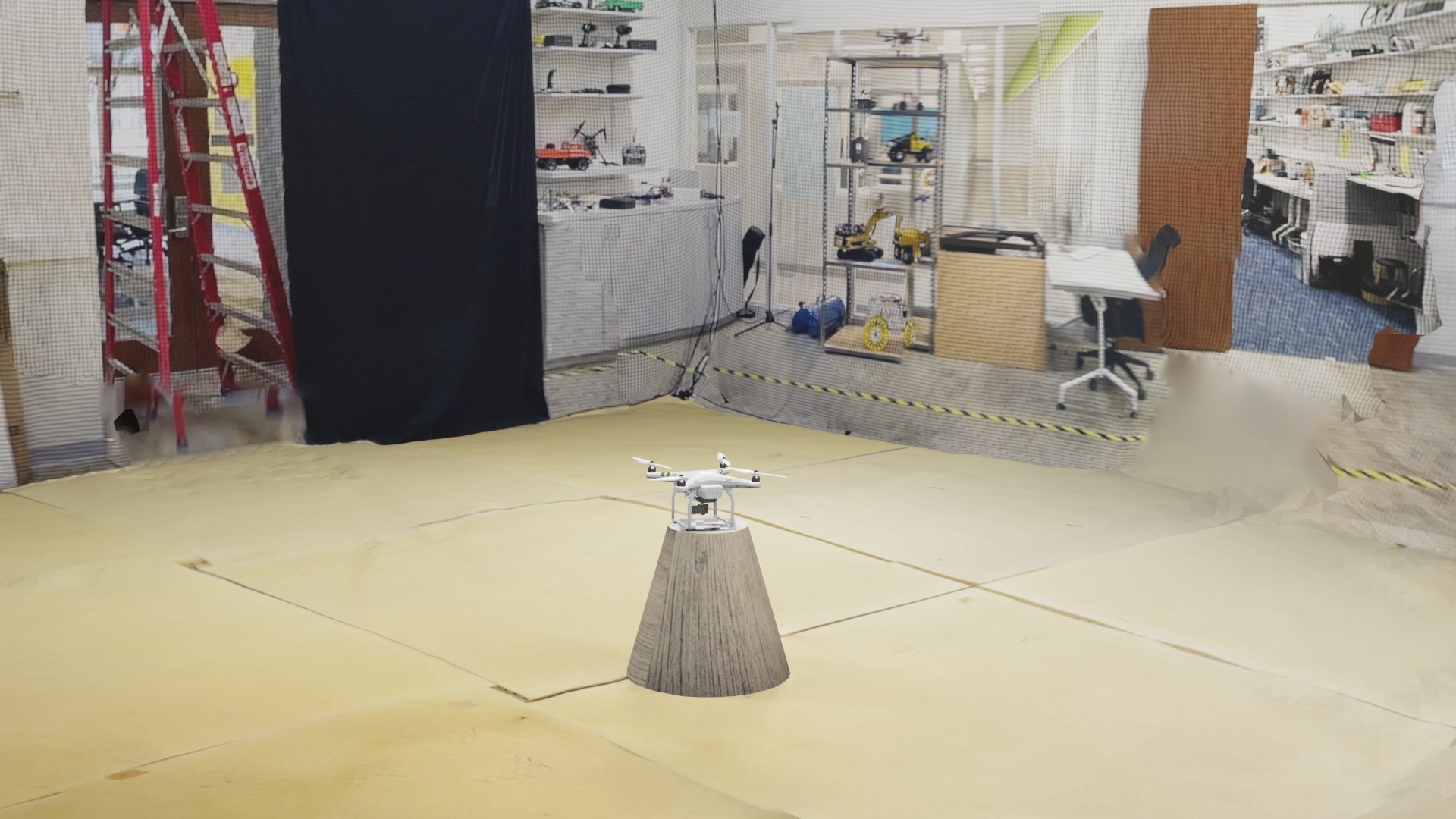}
        \caption{}
    \end{subfigure}

    \caption{Illustration of the safe landing algorithm. The green ellipsoids are the output of the IPC. (a) The uncertainty of the measurement is above the threshold. In this case, the quadcopter will follow the green line and navigate to a new waypoint (blue). (b) The uncertainty of the measurement is below the threshold. In this case, the quadcopter first moves horizontally and then descends to complete the landing. (c) The quadcopter safely lands on the landing pad.}
    \label{fig:illustration_fsm}
\end{figure*}

In order to demonstrate the IPC's capability of being used by downstream modules, we design a simple algorithm that utilizes the output of the IPC to solve the safe landing problem. The proposed algorithm is a state machine as shown in Figure~\ref{fig:fsm}. An overview of the state machine is as follows. In the beginning, the quadcopter estimates the position of the landing pad using its perception module. Then, it runs the IPC to compute the ellipsoid that contains the ground-truth position. If the size of the ellipsoid is above a threshold, the quadcopter navigates to a new waypoint determined by the ellipsoid and returns to the initial state to do another measurement. The quadcopter keeps doing this until the size of the ellipsoid is below the threshold. In that case, the quadcopter will navigate to the center of the ellipsoid and then turn off the motor. We illustrated the algorithm in Figure~\ref{fig:illustration_fsm}. Next, we elaborate on each state in the state machine.

\paragraph{\texttt{Measuring}} In this state, the quadcopter obtains an image from the camera and feeds it to the perception module. The perception module estimates the position of the landing pad as $\hat{y}$. Then, the quadcopter calls the IPC with its current state $x$ and the perceived value $\hat{y}$. The IPC outputs an ellipsoid $\pa(x, \hat{y})$. Intuitively, the size of the ellipsoid reflects the uncertainty in the measurement. If the ellipsoid is small, then the quadcopter can trust the measurements. Otherwise, it has to take another measurement. This is formalized as follows. Let $p_{xmin}$, $p_{xmax}$, $p_{ymin}$, $p_{ymax}$, $p_{zmin}$, and $p_{zmax}$ be the minimum or maximum along each dimension of the ellipsoid $\pa(x, \hat{y})$. If $p_{xmax} - p_{xmin} \leq 0.1$, $p_{ymax} - p_{ymin} \leq 0.1$, and $p_{zmax} - p_{zmin} \leq 0.05$, i.e., the ellipsoid can be completely contained in the box $G$, then quadcopter trusts the measurement and switches to the \texttt{Landing} state. Otherwise, it enters the \texttt{New Waypoint} state and navigates to somewhere else in order to take another measurement.

\paragraph{\texttt{New Waypoint}}
The quadcopter enters this state because the size of the ellipsoid $\pa(x, \hat{y})$ is beyond the threshold.
In this state, the quadcopter will navigate to another waypoint to take a new measurement. The new waypoint is selected as follows.
Since the ground-truth position of the landing pad can be at any point in the ellipsoid, to avoid hitting the landing pad, the quadcopter sets the next waypoint as the intersection point between the surface of the ellipsoid and the line segment connecting its current position $[p_x, p_y, p_z]$ and the center of $\pa(x, \hat{y})$.

\paragraph{\texttt{Landing}}
In this state, the quadcopter will navigate to the point $P_{\mathit{turnoff}} := \left[ \frac{p_{xmin} + p_{xmax}}{2}, \frac{p_{ymin} + p_{ymax}}{2}, p_{zmax} \right]$ and shut off the motors. Recall that the quadcopter enters this state because the ellipsoid $\pa(x, \hat{y})$ can be completely contained in $G$. Therefore, we immediately have the following proposition.
\begin{proposition}
If the ground truth $y$ is in the ellipsoid $\pa(x, \hat{y})$, then $P_{\mathit {turnoff}} \in \{y\} \oplus G$.
\label{prop:contain}
\end{proposition}
Combining Theorem~\ref{thm:pac_bound} and Proposition~\ref{prop:contain}, it follows that with a high probability, the quadcopter will shutoff its motors only inside the target region defined by $\{y\} \oplus G$.

\begin{figure}[tbp]
    \centering
    \includegraphics[width=0.8\linewidth]{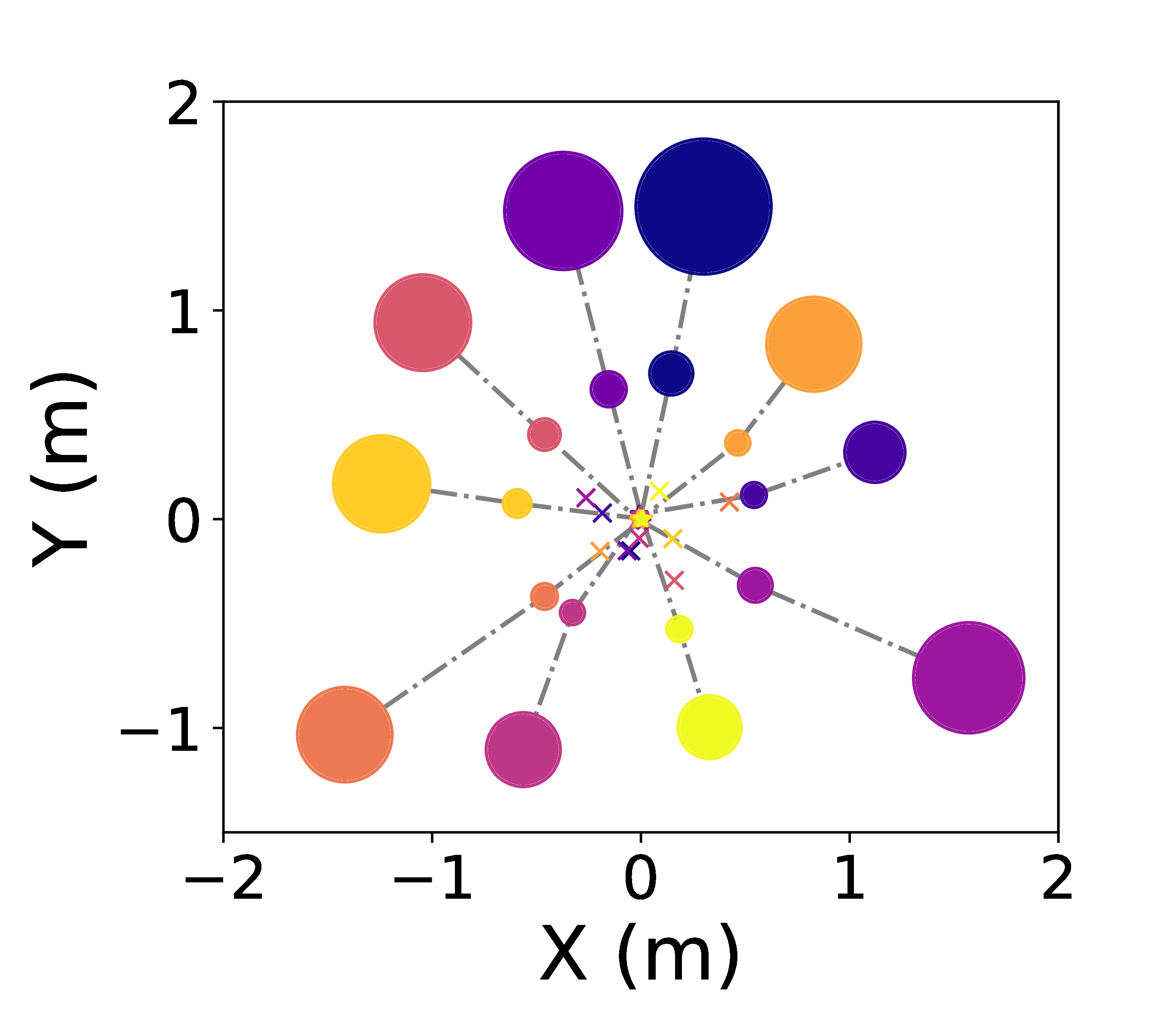}
    \caption{2D visualization of 10 runs. Each run is visualized in a unique color. The crosses mark the positions where the baseline approach turns off the motors. The stars mark the locations where the approach using the learned IPC turns off the motors. Circles correspond to the measurements made by the proposed approach. The center of the circle is the location where measurement is made, and size of the circle corresponds to the size of the ellipsoid computed by IPC.}
    \label{fig:baseline_vs_ours_trajs}
\end{figure}

In order to evaluate the above algorithm, we put the landing pad at $10$ randomly selected locations and ran the above algorithm to land the quadcopter. Furthermore, in order to show the difficulty of the landing problem and emphasize the advantage of the IPC, we also compare it with a simple baseline approach, where we adopt exactly the same state machine as in Figure~\ref{fig:fsm} but replace the learned IPC with a trivial one, namely, $\pa(x, \hat{y}) = \ellipsoid{\hat{y}, 10000 I_3}$, i.e., it always outputs a very small ellipsoid around $\hat{y}$.
The results show that, with the learned IPC, the quadcopter safely landed on the landing pad in all of the $10$ runs, while the baseline approach using the trivial IPC failed in $9$ of them. In Figure~\ref{fig:baseline_vs_ours_trajs}, we visualized the $10$ runs. For each run, we translated the ground-truth position of the landing pad to the origin. In order to make a clear visualization without occlusion, we also rotated the data of some runs. As can be seen from the figure, due to the existence of the perception error, the baseline approach using the trivial IPC landed the quadcopter at locations far off the origin, while with the learned IPC, the control algorithm refused to trust the first measurement and conduct another more accurate measurement, which leads the quadcopter to the precise location of the landing pad. Videos are provided in the supplementary materials.

\section{Conclusion}
We demonstrated that learned inverse perception contracts can be easily used to achieve reliable and safe control.
However, limitations and research opportunities still remain. In this paper, the control algorithm is straightforward and mainly used to demonstrate the perception contract. It does not have any convergence guarantee, i.e., new measurements are not necessarily more accurate. In future work, we will explore more sophisticated algorithms to jointly synthesize the perception contract and downstream controller.

\bibliographystyle{IEEEtran}
\bibliography{IEEEabrv,references}
\end{document}